# A COMPREHENSIVE REVIEW OF PAST AND PRESENT VISION-BASED TECHNIQUES FOR GAIT RECOGNITION

Tracey K. M. Lee, e-mail: tlee@sp.edu.sg
School of EEE, Singapore Polytechnic

Mohammed Belkhatir, e-mail: mohammed.belkhatir@univ-lyon1.fr, phone: +33472692194
Faculty of Computer Science, University of Lyon, France

Saeid Sanei, e-mail: s.sanei@surrey.ac.uk, phone: +441483684615
Department of Computing, University of Surrey

---

Global security concerns have raised a proliferation of video surveillance devices. Intelligent surveillance systems seek to discover possible threats automatically and raise alerts. Being able to identify the surveyed object can help determine its threat level. The current generation of devices provide digital video data to be analysed for time varying features to assist in the identification process. Commonly, people queue up to access a facility and approach a video camera in full frontal view. In this environment, a variety of biometrics are available - for example, gait which includes temporal features like stride period. Gait can be measured unobtrusively at a distance. The video data will also include face features, which are short-range biometrics. In this way, one can combine biometrics naturally using one set of data. In this paper we survey current techniques of gait recognition and modelling with the environment in which the research was conducted. We also discuss in detail the issues arising from deriving gait data, such as perspective and occlusion effects, together with the associated computer vision challenges of reliable tracking of human movement. Then, after highlighting these issues and challenges related to gait processing, we proceed to discuss the frameworks combining gait with other biometrics. We then provide motivations for a novel paradigm in biometrics-based human recognition, i.e. the use of the fronto-normal view of gait as a far-range biometrics combined with biometrics operating at a near distance.

---

## 1. Introduction

The use of biometrics has caught the imagination of the public eye. Using a measurable characteristic of a human, it has become feasible to use automated means of authentication.

In a certain sense this is not new, as older “analogue” recognition technologies have been used for a long time. Thus we had handwritten signatures, thumbprints. What is different now is the use of computer technology to measure the feature. Some examples are: face and facial features, fingerprints, hand geometry, handwriting, iris, retina, vein patterns, voice and so on.

Another consideration is that most of the biometrics used now require that the subject be physically close before the authentication system can work. Close-distance biometrics, for example fingerprint and iris have high accuracy in identification rates, exceeding 99.99%. It is also becoming apparent that in large scale deployment of biometric systems, one biometric is not enough [1]. As of writing, in systems currently deployed, at least two biometrics are used: fingerprint and either face or iris. This aspect will be briefly covered in a later section.

A further reason for multi-biometrics systems is for faster processing. In checking identities, the dataset to be checked is usually large in size. In order to reduce the amount of computation, it is expedient to narrow down the space of possible identities of the subject in the recognition process. Thus the use of other biometrics which operate at a further distance may be employed. So for each biometric, or combination thereof, current research aims to find more robust and quicker means of obtaining the required information.

In this rapidly developing field, one approach is to focus on more subtle biometrics like vein patterns, hand shape, ear shapes and infra-red heat signatures. For existing biometrics, there is still a lot of research being done on face recognition for example, using tridimensional approaches and using other features. On the other hand, concerning biometrics which can be measured from afar, gait promises to be an interesting field.

With such aphorisms like "what you say reflects who you are", the way we walk is unique and can be considered as an identifying trait. Unlike several other biometrics, gait has desirable properties:

- it can operate at a distance of 10 meters or more
- it is non-intrusive as it does not require the cooperation of the subject
- it is non-invasive as it does not force the subject to behave in a certain way
- it is hard to disguise: it has been shown that gait does not vary much in an individual unless there is a case of extreme physical change like carrying a load or change in footwear
- it can be used in infra-red image sequences

In the context of the use of video cameras in a surveillance mode, gait is characterized for identification purposes and can be measured from a distance, far before a face can be clearly seen. Thus it can be used to pre-select a group of possible candidates to check. When the person draws near enough for face recognition to take place, the number of people to check can be drastically reduced, thus giving an overall better recognition rate.

Thus a promising area in recognition is the fusing of several biometrics to make the overall process more robust. The expert knowledge embedded in each system can be brought to bear to make classification more robust. It makes sense to use as much available information as possible to perform the task of identification.

A useful demarcation between the types of Gait Analyses is in order here. The study of human gait was traditionally in medical studies which were diagnostic in nature. It is a part of the field of biomechanics and kinesiology and related to other medical fields like podiatry. The gait metrics are derived from sensors attached to a human and in some cases, video sequences of gait are captured. In this respect, we may term this as Clinical Gait Analysis. This field of study has provided much of the terminology used in gait analysis as well as the initial experiments on recognition, e.g. the various phases of a walk and the various parts of the body used in walking.

In the context of the use of video cameras in a surveillance mode, gait is characterized for identification purposes. Notably, an intrusive process is not suitable. This calls for a fully automated method of tracking and analysis. For lack of a better term, we call this Video Gait Analysis and will be the main context of the term Gait Analysis.

Of all the possible approaches to human motion analysis, only a small subset has been employed for the task of human recognition by gait. This is attributable to the current limitations of technology, namely relatively low resolution video capture devices and the compute intensive task of recovering the pose of a person given a stream of bidimensional images such as in video. To compound this, surveillance videos are taken at a distance so the captured image of a human is even smaller. However with the advent of new technologies it may be possible to deploy more powerful techniques for analysis and with this in view we review some promising and current approaches to general human motion analysis with a view to possible future applications to gait analysis.

In the remainder, we detail state-of-the-art techniques for human recognition by gait followed by some approaches in modeling human gait in section 2. At the end of this section, we give a critique of the main approaches to gait recognition and a comparison between them. We then deal in section 3 with issues and challenges related to gait processing, namely gait tracking, looming adjustment and occlusion compensation. We analyze the combination of gait with other biometrics for human recognition in section 4 and discuss experimental issues related to existing gait datasets in section 5. We motivate for the use of the fronto-normal view of gait in section 6. We finally conclude this paper by summarizing our analysis and offering perspectives for the application of gait analysis and modeling in surveillance frameworks.

## 2. Gait modeling and recognition

To give some background and provide the motivation for identification by gait, we look at early experiments in this field which actually came from psychophysical studies. We start off by considering work done on tracking humans and recognizing walking motions. This field of research is very wide and diverse, going into the fields of virtual reality and Human Computer Interface. However because gait is a specialized human motion, features and algorithms used in motion recognition may be used in gait as well.

We do not consider work done on recognizing between different human activities, i.e. differing gestures as in sign language and the like. This is because these actions can be quite different, and thus features derived from these activities can be distinguished quite readily.

However, to differentiate between various gait patterns is more difficult as generally the activity of walking involves motion that is quite uniform among humans.

This is the focus of our paper. Then we consider the two main approaches to gait recognition, namely model free and model based ones. The definitive early comprehensive survey on human motion analysis was done by Aggarwal and Cai [130]. Although dated, their taxonomy of human motion analysis have been used in many papers. A more recent survey was compiled by Moeslund et al. [2].

### 2.1. Psychophysical considerations

Bio-mechanical and clinical studies in the 1960's show that the actions of hundreds of limbs, joints and muscles combine to give each person a unique way of walking. Not only that, gait can reveal the presence of certain sicknesses, moods and can distinguish between genders. It was shown that variability of gait for a person was fairly consistent and not easily changed, while allowing for differentiation with others. In these early studies, there were hints about the notion of recognition by gait, but the research was mainly medically oriented.

Johansson's experiments with Moving Light Displays (MLDs) [3] gave the “most vivid impression of a person walking”. He showed that ten lights were enough for identification of human motion. These displays comprised reflective tape attached to points of movements. At all times, the silhouette of the body was not used. This psychophysical study showed that a static jumble of lights makes no sense to a user, but when walking, the cadence and position of the MLDs allowed for identification of the subject as a human. For fronto-parallel (side profile) motion, the ten lights comprise: two each of the elbows, wrists, knees and ankles and one each of the hip and shoulder. For more general motion, two more lights are used at the shoulder and hip level.

Furthermore, Johansson showed that the bidimensional motion can be separated by vector analysis. Motions common to all MLDs can be decomposed into essentially translatory, rotatory or pendulum motions. Later on, the common motions, for example, hip and shoulder motion were either removed or modified. But still human motion was recognizable. As with earlier theories on vision, the grouping of the lights give a *gestalt* effect that is learnt by human experience, namely that the total effect of the moving lights is greater than the sum of effects of the individual lights. In the same paper, he reports in his Demonstration 2, that angles from 45 to 80 degrees, can also give the impression of a walking person, implying the view-invariant nature of motion recognition. However, whether these motions can distinguish *between* humans remains to be seen.

Cutting and Kozlowski [4], in follow up experiments using MLDs, confirm that the recognition process was direct, although observers were able to explain the features that guided their choices. Very little training was required to get good recognition. As compared to just looking for the presence, they show that it is possible to identify a person from MLDs.

With other collaborators, they work on attempting to identify between genders. Looking at features like shoulder/hip size ratios, torsion of the torso among others [5]. They come up with the center of moment, a point constructed from various limb measurements. The limbs are treated as pendula attached to the torso which is modeled as a spring. This feature accounted for 75% of the variability in gait. Finally, they postulate the sequence of events that occur when recognition due to gait takes place, starting from the center of moment of the subject [6]. They come up with a computer program to synthesize gait using the center of moment. The graphical output was able to simulate the results obtained so far with human subjects [7]. Yu et al. [8] conduct psychology experiments on gait-based gender classification, and proposed an automated approach based on the weighted sum of block features corresponding to five parts of a body silhouette: head, chest, back, hip and leg.

Little and Boyd [9] conjecture that since the frequency of all the moving points of a human are the same, it does not matter where the MLDs are. They conclude that humans do not derive structure information when viewing MLDs. However, tests have shown that when the MLD is inverted, it is not perceived as a human walking.

The question is: what mechanism does the human mind use for this task? Does it first reconstruct a tridimensional picture from which identification takes place? If so, this is essentially a Shape-from-Motion (SfM) problem. However, from face recognition work using eigenvectors, we see that it is possible that features based on the training set itself can be used for recognition, without any prior knowledge of human features. This is the motivation for model free approaches, which use only the gross features of motion for identification purposes.

Because of the breadth of topics covered, we provide a graphical overview of this paper in Fig. 1. The actual flow of concepts is that from a biometric viewpoint and thus is covered in the earlier sections, so in section 2, we give an overview of the methods used in papers which deal specifically with gait recognition. We survey model free methods and in considering model based methods we also look at various approaches to modeling gait, which are not currently used for recognition, but which can provide useful features for future recognition approaches.

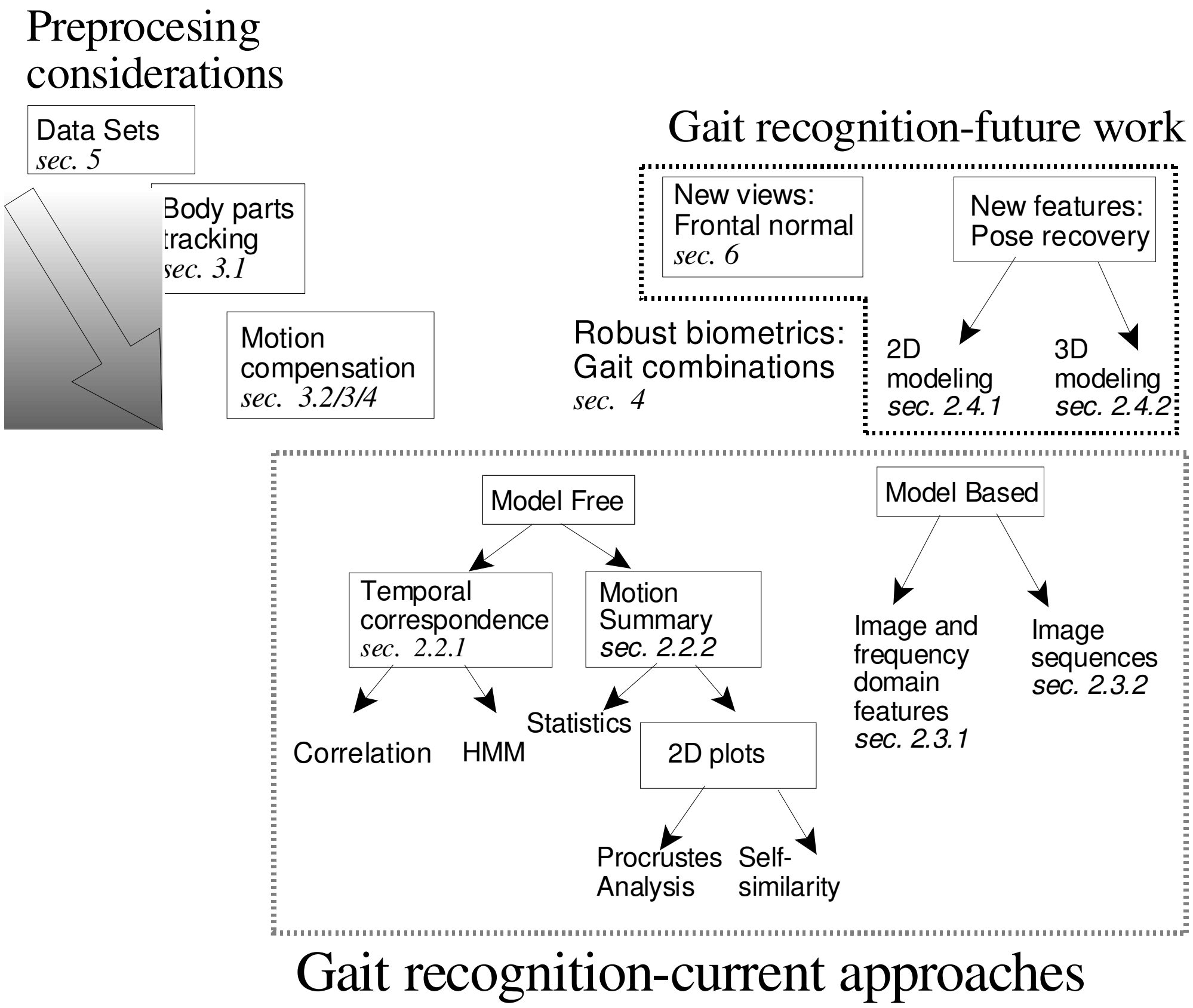

Fig. 1 Overview of gait recognition approaches

## 2.2. Model free methods

In this set of approaches, gait can be considered to be made up of a static component based on the size and shape of a person and a dynamic component which reflects the actual movement. In gait recognition the static features are height, stride length and silhouette bounding box lengths. Dynamic features used are frequency domain parameters like frequency and phase of the movements.

The computationally expensive and detailed tasks of model reconstruction give way instead to recognizing just the gross movements of a walking person. This is normally done by obtaining the silhouette of the person to isolate the distracting background. This has been the mainstay of gait recognition approaches to date.

We look at instances where gait specifically is used for identification. We concentrate on features and pattern recognition approaches. There are two main approaches here, namely temporal comparison on a frame by frame basis and summary of spatio-temporal information. Since the testing methodology for gait was standardized recently and there are several datasets available, it is not useful to compare results *across* papers so we will only mention the results of an approach if it is significant. The name and references of the datasets are mentioned in Section 5.

### *2.2.1. Temporal correspondence*

This is done by comparing and matching features between the test input and the existing patterns; i.e. spatial comparison is performed temporally, on a frame by frame basis. These approaches use the image data directly, comparing the test image sequences with the training sets. In this case, the main pattern recognition approaches use correlation and Hidden Markov Models.

A significant milestone in gait recognition has been the setting up of a test dataset by Sarkar et al. [10] as part of the HumanID project. A rather realistic setting was achieved by videotaping the subjects in sunlight with shadow. There were standard variations of the walker, including differences in view, type of footwear and walking surface. A manual method was used to define the bounding boxes of the walker. Then standard algorithms for background subtracting were used to reduce noise and derive a binary silhouette of the walker with dilation of pixels. This silhouette was then normalized in size. The foreground subtraction routine was statistically based. A *baseline* gait recognition algorithm was provided which allowed a *probe* walker sequence which was a stride cycle, to be compared to those in the *gallery* (database) sequences. This used a correlation between frames, using the ratio of the *intersection* to the *union* between the probe and gallery frames. A table of results shows that even a simple algorithm provided encouraging recognition rates, in the 70-80% range for "simple" tasks, dropping to 10-30% for "difficult" tasks.

Some interesting results [10] were that the lower 20% of the silhouette, that is, from the knee down, provided for 80% of the recognition. Also, changes in surface type caused changes in walking patterns by up to five times as compared to those caused by differences in shoe type. Viewpoint did not play that much a part, presumably because the image capture was done at a distance which allowed an approximation to a fronto-parallel walk to the camera. This is considered a correlation approach.

Kale et al. [11] came up with a novel method to derive the canonical pose for walking and apply it to a simple version of the baseline algorithm as provided by Sarkar et al. [10]. This pose is the fronto-parallel one. After showing that perspective, rather than orthographic projection is effective here, they use optical flow methods to track a feature on the walker and thereby derive the angle of walk. With an assumption of walking along a straight line, they find this angle and warp the image so it is fronto-parallel to the camera. By fusing the height information with the leg dynamics, a better recognition rate was achieved. Walking at an angle of 45 degrees still gave encouraging results.

Kale et al. [12] use Hidden Markov Models to distinguish between temporal data. They work on a binary silhouette obtained by background subtraction and noise erosion. The feature used is the *width vector*, which is the difference between the left and right edges of the silhouette boundary and has as many elements as the number of rows of the image. From analysis of width vector profiles [13], only one half of a walk cycle is sufficient and provides static *and* dynamic information about the walker. To be of practical use, this vector needs to have about 100 elements. To compute statistics for a vector of this size requires about 5000 training samples, which is not practical. To improve on this situation, several gait cycles of a walker are examined. Using k-means clustering, five images are selected from a video sequence. These images are the most representative of the stages a walker goes through in a gait cycle and are designated as *stances.* When a test image is obtained, a new feature vector is formed by taking the L1 norm of the width vector with each of the five stances, thus reducing the number of dimensions to just five. To perform identification, the sequences of images from a walker are taken and the likelihood it was generated from the Hidden Markov Model corresponding to a particular person is computed. The highest likelihood is the most probable model then. Further experiments show that the 5-state Hidden Markov Model gives the best results as compared to 3 and 8-state models. This method was sensitive to changes in the viewing angle.

Ran et al. [14] present two methods for estimating the period of gait cycles. The first is based on a Fourier transform and a periodogram while the second uses Maximal Principal Gait Angle (MPGA) fitting. Both methods are used for pedestrian detection. Cycle characteristics are determined based on the detection of the phase difference between the input and output signals of a Voltage Controlled Oscillator (VCO). Ho et al. [15] estimate gait cycle through both static (i.e. the highlighting of motion vector histograms) and dynamic (i.e. the extraction of Fourier descriptors) information. Dimensionality reduction of the used features is achieved through Principal Component Analysis and Multiple Discriminant Analysis. A nearest neighbor classifier is then used for gait identification.

#### *2.2.2. Towards Spatio-temporal Motion Summary*

In an effort to reduce the challenges of comparing images on a frame by frame basis, it is more efficient to use summarized motion features. The video signal is a bidimensional image which changes over time. A typical walking cycle may be summarized spatially, temporally or both. With reference to Fig. 1, this can be done by obtaining the statistics of motion detected in the images in a walking cycle. In a similar way, a spatial summary may be obtained by converting the bidimensional silhouette into a single quantity which can be analyzed to derive time varying features which can be used for identification. For example, Hu et al. [16] propose the use of Gabor filters to represent body shapes through several orientations and scales. Dimensionality of the used features is reduced through the combination of Principal Component Analysis and Maximization of Mutual Information. Gaussian Mixture Models and Hidden Markov Models taking gender into account are trained for gait classification. The authors report experimental results on par with the state-of-the-art. Venkat et al. [17] partition a silhouette into overlapping upper, middle, lower, left and right parts and use a Bayesian network for gait identification.

Now, by summarizing the motion temporally as well, an image can be formed by the moving average of the value of pixels which have motion in a video frame. Thus a single *motion template* figure can be used for recognition using standard image comparison methods. A fixed average image may also be used. Similar motion template figures are grouped and pattern recognition is performed using methods like e.g. k-nearest neighbor.

Lee [18] works on the statistics of the movement of the silhouette of a person walking perpendicular to the image plane. The image of the walker is transformed into this image plane, making this a *view* and *appearance* based approach. She divides the binary level silhouette into seven elliptically shaped regions. Geometric measurements of these ellipses form the recognition features. This also introduces some noise tolerance. It was considered to use the period of variation of those features. But attempting to find the periodicity of these features was difficult at lower frame rates. However features like the amplitude of leg, arm swing and head orientation are more useful. This can be represented by the mean and standard deviation of the features. Thus the period is not used and the final features are the mean and standard deviations of the parameters of the ellipses for a given sequence of motion. In this way, the summary of the motion is computed.

BenAbdelkader et al. [19] track the subject in a video stream and extract the bounding box of the extremes of motion. A series of such boxes is generated from the video frames at different times and normalized to a common size. These scaled boxes, known as templates, are subtracted from each other pixel by pixel (spatially) for all given pairs of time instances (temporally). Then Cutler and Davis' two-dimensional *similarity plot* [20] is generated by plotting the sum of all the image pixel differences between frames for all time instances. This plot then is a summary of motion and is subjected to an eigenvector type of analysis for pattern recognition. Hence this has been called the eigengait approach.

Wang et al. [21] consider the dynamic and static figures of a walker. They use the “background primal sketch” [22] to segment and build a sequence of silhouette images. The bidimensional silhouette is converted into a unidimensional distance signal based on the work of Fujiyoshi and Lipton [23]. The silhouettes are “unwrapped” using the geometric centroid of the silhouette as the origin of the image. A set of points along the silhouette is selected by sampling the silhouette at a fixed angular distance with respect to its origin as shown by the lines in Fig. 2. The *sum* of the Euclidean distances of these points on the silhouette *from* the origin (i.e. the length of the lines in Fig. 2), form a time-varying unidimensional signal from the bidimensional silhouette image. This distance is normalized by magnitude. A set of distance signals will thus be obtained from a walking sequence. All the distance signals from all of the walkers are subjected to eigenvector decomposition to reduce the dimension of the problem. Building on the work of Murase and Sakai [24], they also keep time stretched versions of the distance signal. A Nearest Neighbour Comparison is done on this dynamic feature of a walker for recognition. Later they incorporate static features like height, speed, maximum and minimum aspect ratios of the silhouettes. By combining static and dynamic features, they are able to obtain a 100% success recognition rate on the Soton dataset.

Wang et al. [25] extract the moving walker, yielding a set of unwrapped silhouettes described earlier in Sec. 2.2.1. These silhouettes are subjected to Procrustes shape analysis. Three views are used, being 0, 45 and 90 degrees to the camera. An eigenvector analysis yields the mean image for this cycle. This is thus a spatio-temporal summary of the walk. The mean image is used to compare among different walkers. Using a nearest neighbour approach, they report an 88 to 90% recognition rate.

Lu et al. [26] employ three kinds of analyses on three views of unwrapped silhouettes to derive features. This is followed by Independent Component Analysis to reduce the data dimension. Then they use three types of classification methods on the resulting data.

<u>Motion History Image and Variants</u>

A significant way of obtaining features from a video sequence involves summarizing the motion spatiotemporally into a Motion History Image (MHI) as first described by by Davies and Bobick [27]. In fact, the importance of this method and its variants has been reported by [28]. The MHI is generated by identifying *moving* pixels in successive frames of the motion. The MHI describes *how* the motion proceeds by assigning a value to these pixels. In successive frames, *previous* moving pixels have their values decreased. Thus the most recent frames contribute brighter pixels to the MHI. The associated Motion Energy Image (MEI) identifies *where* the action is taking place by performing a logical OR on the pixels between the frames (an example is shown in Fig. 3). Thus a bidimensional image now represents the motion and standard pattern recognition methods based on *static images* may now be used to identify people. Based on this, Han and Bhanu [29] use Gait Energy Images (GEI) to identify persons by their gait.

Zhang et al. [30] improve on GEI with Active Energy Images (AEI) that concentrate on the actively moving parts of a gait silhouette, reducing the effect of movements just due to translation. They also apply two-dimensional Locality Preserving Projections (2DLPP) to the AEI for better pattern discrimination. Silhouette images are inherently noisy and Chen et al. [31] devise the Frame Difference Energy Image to compensate for incomplete silhouettes. Missing information is derived from silhouettes in other frames which are clustered according to their GEI. The results show promise in the use of this technique.

Chen et al. [32] propose to extract a representation based on Gabor features of a GEI on which they apply a Riemannian manifold distance-approximating projection called TRIMAP. The latter is based on the construction of a graph from data preserving pairwise geodesic distances and then optimizing the discrimination ability. The reported gait recognition results outperform those of a baseline implementing Latent Discrimination Analysis.

An interesting variant is the use of Moving Motion Silhouette Images (MMSI) by Nizami et al. [33] which summarize the motion of *subsets* of the silhouettes obtained for a complete gait sequence. Thus there are a few MMSI for a gait sequence and these are subjected to ICA as a dimensionality reduction measure. The resulting Independent Components derived from *each* MMSI are classified using probabilistic Support Vector Machines. The results of each SVM contribute to an Accumulated Posterior Probability and are fused to give an overall identification result. The authors quote a high identification rate of 100% for the NLPR (or CASIA A) dataset and 98.67% on the SotonBig dataset for their proposed methods.

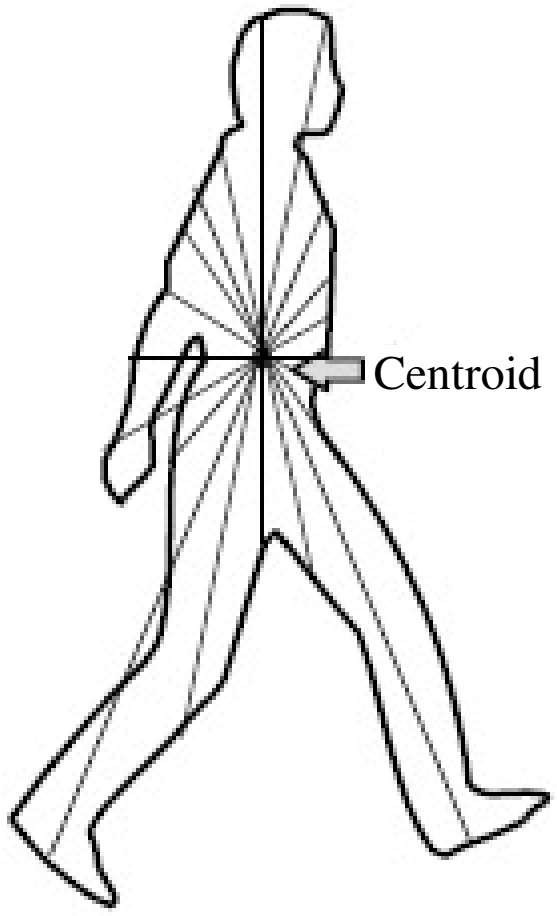

Fig. 2. Unwrapping a FP view silhouette – distance from points on silhouette to centroid

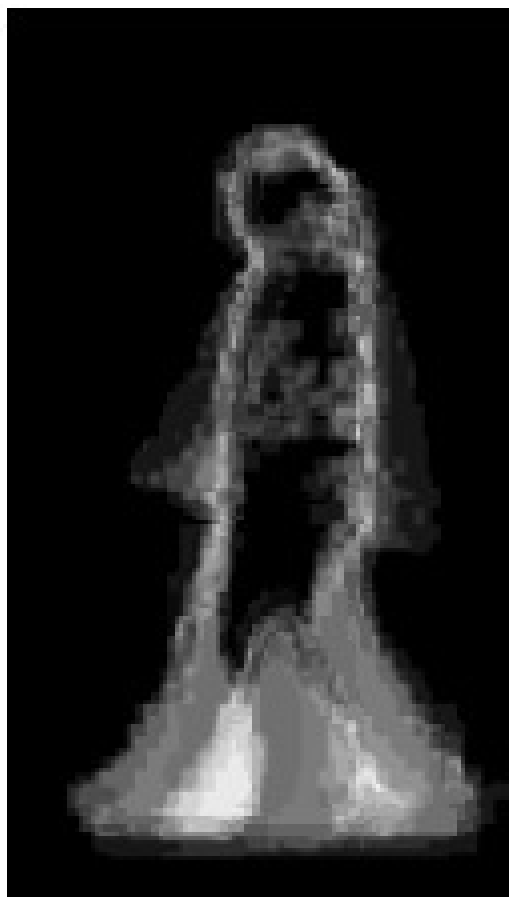

Fig. 3. MEI of a person, brighter areas represent more movement

## 2.3. Model based approaches

The premise for the model based approach is based on the fact that interpreting the basic features in an image such as edges and lines require prior knowledge about the image to speed up the process of recognition. This knowledge comes in the form of a model which can cope with ambiguities of interpretation and guide the search for movement parameters in a directed way as we examine approaches for human recognition by gait. One main set of approaches use waveform derived features such as frequency and amplitudes together with image features while another directly use a set of images directly so that time is an implicit feature.

### *2.3.1. On the use of static and dynamic features in model based approaches for recognition*

Gait can be considered to be made up of a static component based on the size and shape of a person and a dynamic component which reflects the actual movement. In gait recognition the static features are height, stride length and silhouette bounding box lengths. Dynamic features used are frequency domain parameters like frequency and phase of the movements as exemplified by Cunado, Nixon and associates. Bobick has used both static and dynamic features such as trajectories of joint angles.

Cunado [60] and Nixon [61] analyze the motion of the thigh and calf, deriving the frequency and phase of the movement. By using a fast Hough Transform (HT), a walker is represented by a pair of lines in a “lambda” shape. This affords simplification of articulated movement and takes on the HT’s robustness to noise and occlusion. From the HT, the features tracked are the hip movement, modeled by a linear and oscillation term and the angle between the legs, modeled similarly by a Fourier series. Looking at the magnitude of the features, there is little variation among subjects and while phase gives better differentiation, the corresponding magnitude is small. By multiplying both the magnitude and phase, the “phase-weighted” magnitude is a better distinguishing feature. The authors use k-means clustering for classification with encouraging results on a small dataset.

As a later development, Foster et al. [62] attempt to introduce a degree of model specific information in the form of masks that measure the area change of a human silhouette. These masks are formed by detecting the changes in the image of a moving person and are specific to humans. By using Canonical Analysis on the AC (Alternating Current - to denote the changing, not static) component of the output of the masks, they report a 81% success rate on recognition.

BenAbdelkader et al. [63] use stride length and cadence (walking frequency) to characterize a subject. By segmenting the walking object and constructing a bounding box around it, they are able to derive the frequency of walking by calculating the change in size of the bounding box from the video frames. The pose of the walker is found to affect the frequency of gait but could be constrained by noting the range of human gait. Later they follow up by including height as another biometric, obtained from image measurements [64]. However, they use the fixed and variable part of height (caused by walking) and show that the identification performance improves from 51% to 65%. They use first bidimensional then 4-dimensional feature vectors for pattern recognition. These papers give an extended discussion on measurement errors as the images are derived from low resolution video.

Bobick and Johnson [65] look at four *static* parameters of a person’s walk, namely the bounding box of the walker’s silhouette, the distance between the head and pelvis, the maximum distances between the pelvis and left/right foot and the distance between the right and left foot. In order to account for the variation in viewing angles, each of the parameters is scaled by a view angle dependent constant. The scale factors are measured using magnetic sensors on “reference” subjects in a separate experiment. The binary silhouette of the walker is derived by background subtraction. By grouping the pixels into five body regions, the effect of noise is minimized by using the centroids of these regions. Instead of using standard error measurements, the authors use the measure of “expected confusion”, that is how much the measurement reduces the uncertainty of identification. This measure aims to give a better idea of how the identification will perform in larger datasets. In a later development [66], the authors discuss how shadows are removed using Gaussian modeling of the image pixels, for shadow and non-shadow regions of the image. The general locations of these regions are estimated using the camera location relative to the sun.

Tanawongsuwan and Bobick [67], instead of static, investigate the use of *dynamic* walk characteristics (i.e. joint angle trajectories) as recognition features. They use the angles of the left and right hip and knee joints. Being an initial investigation, they use magnetic sensors attached to the subjects. The placement of such sensors could affect the results. The joint angles are calculated with care using the structure of the underlying human skeleton, so as to reduce accumulation of errors. Since they are looking at dynamic characteristics, normalizing the trajectories is needed. First, the signals are normalized so that they have unit variance, then they perform Dynamic Time Warping (DTW). These signals are then time aligned to be of the same duration and general shape - for each trajectory of interest. The feature used is a 240 dimension vector of the four sampled trajectories concatenated. This is then reduced to 4 dimensions by Principal Component Analysis.

#### *2.3.2. Temporal Representation & Modeling*

In the temporal representation approaches, time is assumed as the third dimension complementing the XY axes in the image plane. A gait sequence is therefore represented in a XYT three-dimensional space.

Niyogi and Adelson [53] use a XYT approach where the images from a walking sequence (XY) are stacked up to give an image cube where the third dimension is time. By analyzing the XT plane, the walker's ankle traces out a unique braided pattern as compared to the head which traces out a line. This is used to detect the presence of a walker in the image sequence. The width of the braided pattern is determined by the edges of the body and limbs as they move. The edges of the braid are found by using active contour models or "snakes". The snakes are tuned to account for discontinuities at the hips, knees and ankles. The contours are formed at various levels of the XT slices and this gives a silhouette of the walker (in the XY plane). By averaging the silhouette, they create a stick figure representation of the walker. For recognition, they use four joint angles from the stick model. The temporal values are warped to give a common time base. They then use the derivatives of these angles and interpolate them to form a 40 dimension vector. To compare the input image sequences with that in a dataset, an L2 norm was used between all pairs of these sequences. They report a 79% recognition rate for recognition of a person by their gait. In a later work [54], they use the XYT data directly, by fitting a so-called spatiotemporal surface to it. This surface in turn is derived from six parameters, namely the start and end positions of the walker, bounding box size of the image, period of walking and phase. This specifies the canonical surface. Part of the canonical surface is derived from analyzing the walks of all the subjects. By analyzing the actual image data, the deviation from the canonical surface is noted. Now, the deviation may be modeled by applying "forces" which cause the canonical surface to be deformed according to the subject's image data. The authors do not perform any classification, but we see that the parameters describing the canonical and deviation surface would provide enough information for doing so. This approach does not face the problem of occlusion. If there are more than one walker, a Hough Transform may be used in the XT plane to separate them out. Kellokumpu et al. [55] make use of a three-dimensional binary representation in a spatiotemporal space to characterize human walking movements. For multiresolution, they start from this representation for extracting XYT histograms. They show improved results in comparison to state-of-the-art multiresolution analysis on gait recognition using the CMU Mobo database. Ran et al. [56] decompose a gait sequence into XT slices to generate a periodic pattern termed the Double Helical Signature through an iterative curve embedding algorithm. This representation is used to highlight body parts in challenging environments for gait recognition, typically in the context of surveillance (i.e. cluttered scenes and load-carrying conditions).

In the modeling approaches, the authors propose to design models taking into account the temporal dimension. In [57], a dynamical model of human motion based on self-regression functions is used to model the gait cycle. The latter is divided into four approximately equal phases in [58]. The relation between horizontal and vertical accelerations as well as other indicators is derived by a rule-based approach following the Computational Theory of Perceptions. Zhang et al. [59] propose two generative models representing the kinematics and visual appearances of gait through latent variables, namely the Kinematic and the Visual Gait Generative models.

The amount of research of model based gait recognition is much less compared to model free approaches – this is probably due to the computationally intensive task of full scale pose recovery. However it is useful to examine such approaches which derive gait by synthesis of human pose in the next section, so as to consider features which may be used for recognition.

## 2.4. Model based approaches to pose recovery

Recovery of human pose starts with two dimensional video images. In staying with this lower dimension, there is compactness in data representation and manipulation, but one has to bear in mind that one is working with the projection of an actual three dimensional object where there is some loss of information.

### *2.4.1 Bidimensional Modeling*

Marr and Nishihara's [34] early paper on features for the two-dimensional representation of human figures have been influential. They discuss the various ways of representing a shape, from skeletons to a set of jointed cylinders. Later, Marr [35] gave further details from a psychophysical point of view and more detailed account of how to actually implement the ideas discussed.

Ran et al. [36] propose the combination of an edge detection method and Hough Transform to extract the Principle Gait Angle, defined as the angle between two legs during walking. A Bayesian classifier considers the frames of the gait sequence that are categorized according to whether the Principle Gait Angle is positively or negatively detected.

Jean et al. [37] analyze the trajectories of the walker's head and feet. However, issues related to occlusion and looming are to be addressed (cf. Section 3). Another problem is caused by foot movement for which the separation between feet in each frame is computed to determine whether the front foot changes.

<u>Scaled Prismatic Model</u>

A significant representation of two-dimensional motion is the introduction of the Scaled Prismatic Model (SPM) by Morris and Rehg [38]. In order to recover the model giving rise to a two-dimensional image stream, kinematic considerations define a state vector of joint angles as the desired objects to be derived. They also provide the mapping between these states and the two-dimensional image features. The SPM, being a projection of tridimensional objects, appears as a link which can rotate and be scaled. Borrowing from robot manipulator theory, nonlinear least squares methods have been successful in this instance. Now the two-dimensional images have to be differentiable and observable. But in monocular image sequences, singularities occur in certain tridimensional configurations. The tridimensional recovery problem can be split into a registration and reconstruction process. By using the SPM in the registration process, these singularities can be better handled. The reconstruction of the tridimensional model from the SPM is done as a batch process.

Taycher et al. [39] use the SPM for motion recognition by recovering the pose of a body. By considering that the body is made up of a set of rigid segments, this is expressed as a graph, where there are as many nodes as there are segments and the edges have values of zero or one depending on whether they are connected. The nodes represent the pose of a joint of a limb. To find the optimal tree representation, they use the maximum likelihood tree that gives the lowest entropy using the maximum spanning tree algorithm. This gives the best estimate of the figure topology, i.e. the connectivity between the segments. An essential step is to calculate the pairwise mutual information between all body segments. This uses the marginal entropy of, and between each node. In actual tests, segment locations are obtained from markers attached to a subject, and also a synthetic model was used.

In using the SPM for tracking, Cham and Rehg [40] focus on using probabilistic methods for tracking the SPM links. Kalman filtering has traditionally been used to track the state vector. In realistic situations, background clutter, occlusions and complex movements give rise to a state space density function that is multi-modal. The authors contend that tracking using current Monte Carlo based techniques like the Condensation algorithm incurs high computation costs. They then use a novel Multiple Hypothesis Tracking (MHT) approach, focusing on the modes of the probability density function, assuming each mode represents a Gaussian probability density function. In probabilistic tracking, it is necessary to update the probability density functions of the states. Using Bayes' Rule, the prior distributions are obtained by the Kalman filter for each mode. In the ensuing calculations, the modes are represented as Piecewise Gaussians rather than using the Expectation Maximization algorithm to find a suitable overall probability distribution function (PDF). Then likelihood is generated from the SPM using image pixel values. A two-dimensional, 19 degree-of- freedom SPM was initialized interactively and allowed to track a video sequence successfully.

As discussed by DiFranco et al. [41], the object motion captured by two-dimensional images in a video sequence can be considered to be degraded by noise, projection and occlusion. Even though one can recover the SPM for the joints in two dimensions, recovery of the tridimensional motion is inherently ill-posed and regularization needs to be imposed. First are kinematic constraints, which describe the connection between links, constrain the link lengths and restriction of rotation *axes*, where applicable. This is still not enough, and the recovery process needs to be supplemented by joint *angle* constraints and dynamics of movement which favors smooth motion rather than abrupt motions. Finally, interactive tridimensional key frames are specified together with pose. All the image sequence two-dimensional SPM data are expressed as state vectors. Together with the motion constraints, these are all combined into a matrix state equation and solved for the *maximum a posteriori* estimate for the state that describes the motion fully. The results are shown for as a reconstructed synthetic view of a short video clip with complex dance sequences.

Although a bi-dimensional representation of a walker data is desirable due to its compactness, deriving the pose from two-dimensional images is difficult because these images are of fully clothed and fleshed people. To match such images, more complex three-dimensional models are needed.

### 2.4.2. Tridimensional Modeling

The pioneering work in this field was Hogg's [42] WALKER program which works on a series of monocular images derived from video frames. The WALKER program attempts to fit the images to various proposed tridimensional models of a human walking. The proposed models for a current image frame depend on the parameters of the previous image and physical constraints of a body's movement. The model is described by a hierarchical ordering of the 14 cylinders that represent the body parts together with a specification of their movement constraints.

The procedure starts by differencing images to find moving points. Bounding the moving object into a rectangle constrains the position of the model. It uses maximal edge points in each image as relatively invariant features and searches for correspondence to the model. A simple data set showed successful tracking between the model and the image sequence.

Rohr [43] also uses similar tridimensional models. From the images obtained from video sequences, a rectangular bounding box around the moving object is drawn. This box is used to estimate the subject's tridimensional position. Within the box, grey valued *lines* are found using an eigenvector technique. A series of models are generated and the contours arising from the model are matched with the grey valued lines using a geometric approach. The search for the initial model position and pose takes the longest time, but assuming smooth motion after that, it will take less resources to track the subject. The model parameters are updated using a Kalman filter approach.

Wachter and Nagel [44] fit the projected model to the image by using the Iterated Extended Kalman Filter (IEKF). There is a feature vector $\boldsymbol{f}$ which determines the fit between model and image. The differences between points on the model and image are used to correct $\boldsymbol{f}$ iteratively, from one image frame to the next. A Maximum A Posteriori (MAP) estimation is performed on edge gradients, both in the model and image segments. By assuming that projected grey values of *surfaces* stay the same between image frames, region information is included in the update. The authors use ten parameters which are manually initialized. As well, the rates of change of the parameters are used in the update process.

Gavrila and Davis [45] use tapered super quadrics on a 22 degree-of-freedom model of a human. Using a front and side view they derive joint angles as features to match the model to the image. They segment the moving image and back-project it to search for the closest tridimensional model. The pose of the subject is determined by *chamfer matching* of edge contours which is described in detail in their paper. However, many views of a tridimensional model need to be generated.

Bregler [46] finds the motion parameters of ellipsoidal regions in the two-dimensional images of a walker. The images are modeled as scaled orthographic projections of a tridimensional model. This model is made up of tridimensional ellipsoidal body shapes. Using concepts from robotics, the pose and motion of the model is represented by *twists*, rather than Euler angles. Since the various parts of the body are linked, points on the various limbs form a *kinematic chain* which depends on the twists of the limbs with respect to each other, in the same manner as Euler angles. Thus a two-dimensional point on a body segment can be expressed by so-called *exponential maps* which perform the required mapping of coordinates. Now, projecting the tridimensional ellipsoids creates two-dimensional support maps which help determine which pixels belong to the particular body segment. Each two-dimensional image sequence updates the model parameters by solving for the tridimensional equations by iteratively using the Newton-Raphson method. However, initialization has to be done interactively. Using a 19 degree-of-freedom model, they were able to track various video sequences well. From this, some possible features for classification may be found in the motion parameters of the tridimensional body segments.

Leventon and Freeman [47] use a probabilistic approach, using training data from a set of tridimensional motion sequences. Each of these are divided into ten frames which are supposed to be the most representative sub motions of human activity. This constrains the kind of tridimensional motions that generate a given sequence of two-dimensional images. These frames form a feature vector whose dimension is reduced to 50. The tridimensional data is made up of the coordinates of 37 body markers. By considering how the data is projected into a sequence of two-dimensional images, they form an equation, using Bayes' Rule. This equation is a multidimensional Gaussian and is the a posteriori probability that this two-dimensional sequence is generated by a certain set of tridimensional sequences. To test out the algorithm, a tridimensional stick figure is used which specifies what should be the two-dimensional equivalent image, using orthographic projection. The difference between the two is used to minimize an energy function. Manual intervention is also allowed in case the minimization does not proceed smoothly. The optimal set of values describing the tridimensional motion is taken and fitted to a cylindrical model for final proofing.

In a later development, Howe et al. [48] refined the above model in the modeling of the a posteriori equation by using a sum of Gaussians as the a priori probability distribution. To do this, the groups of tridimensional frames used as training data is grouped using k-means clustering and a Gaussian PDF is formed around each cluster. The Expectation Maximization algorithm is used in a Maximal A Posteriori way to find the optimal mix of Gaussians as well as the best subset of tridimensional motions that describe the current two-dimensional motions. Notably, a complete sequence of two-dimensional images is a combination of subset sequences of tridimensional images. For sequences parallel to the projection direction, there is some problem with reconstructing tridimensional data, and heuristics are brought in to help. However, poor lighting, occlusion and lighting changes cause the tracker to fail in certain long sequences.

Ning et al. [49] use a tridimensional articulated body model and use the joint angles and velocity as features. They use the Condensation algorithm to track these features on images parallel to the image plane only. The model has 12 degrees-of-freedom and is made up of tapered cones with a sphere. Matching is done with the two-dimensional image using edge and region information [50]. In keeping with statistical methods, the a priori distribution of the features is determined beforehand by analyzing existing image sequences of walking subjects. Since they are not interested in individual walking sequences, the mean and variance of these features are obtained from all training samples. From there they derive the class conditional distributions of the features. This serves to constrain the values of the features as we know that joint angles have limits to their values. They initialize the tracking by observing the first few frames of the test sequence, isolating the moving portion and constructing an edge map. They then correlate it with the sequences in the reference dataset in order to locate the subject. A Pose Evaluation Function is computed, which is a measure of the posture of the subject at a given time. This is done by computing the degree of mismatch between the edges and regions of the model with the images. They report good tracking, but the calculation time runs into minutes, for each frame.

Moeslund and Granum [51] use another Analysis-by-Synthesis approach, employing phase space to describe the motion of the model. As usual, this space is reduced by kinematic and geometric constraints corresponding to body parts movement and placement. The dimensionality may be reduced even more by explicitly considering the physical structure of the limbs. For example the representation of the shoulder-elbow-hand joint can be modeled by a two-dimensional structure instead of six. The silhouette of the image is used to further reduce the search for the tridimensional model. By using a stick figure with a logical AND operation between this and the two-dimensional images, implausible poses may be further eliminated. Finally by comparing between bounding boxes of the projected tridimensional model and the two-dimensional images, further pose elimination is done. All this allows a reduction in dimension. The authors report good tracking results.

From a sequence of three-dimensional gait data, Gu et al. [52] present a method to extract key points and pose parameters automatically. They then estimate the multiple combinations of joints and movement features that are the height, position and orientation of the body. Hidden Markov Models are used to model both movement and configuration features. Their goal is first to classify actions based on a hierarchical classifier with sum and maximum a posteriori rules. Identities can then be recognized from gait sequences.

## 2.5. Assessments and comparisons

In this section, we give a summary of the various approaches to gait recognition and a brief comparison between them and other considerations.

The movements of a walker can be summarized from prominent features. The latter allow one to perform identification with sparse data. As compared to the problem of generalized motion tracking and analysis, work done specifically on identification by gait is sparse. Distinguishing between similar motions of several people may be more difficult than distinguishing several different motions of one person.

In model free approaches, one has to obtain his own features. Pattern matching is done either through a summary of the walk cycle, or else comparison is done on a frame by frame basis. One of the early approaches to gait recognition compared sequences in images directly, using pixels in a two-dimensional array. The alignment of images in time was done by DTW, a process that cannot be done in real time. Later, gross motions corresponding to sinusoidal movements are used to measure the dynamic characteristics of gait in this plane. Model free recognition methods perform surprisingly well, considering they do not make use of the features of the human body. But as we have seen in eigenvector pattern matching used in face recognition, human-recognizable features may not always be the best to use. The variants of Motion History Images have proved to be popular approaches, employing image recognition techniques on an image representing the spatiotemporal summary motion of a subject. In perspective, intuitively we would expect the variation of signal with time to more accurately reflect the idiosyncratic motion of a person. So a spatial summary of motion with time variation would be expected to be more accurate. But when comparing motion, the problem of aligning actions comes in, and as we have seen, this can only be done offline. However as described earlier, Nizami et. al [33] have employed a hybrid approach, whereby *segments* of motion are compared on a spatiotemporal basis, mitigating to a certain degree the effect of averaging over idiosyncratic motion.

In model based approaches, only a subset of the tridimensional human model is used. Even then, a very simplified human model is assumed. Some static human features are used, as well as dynamic features. An interesting development is the Scaled Prismatic Model which may allow probabilistic gait recognition on a two-dimensional level. Most work on model-based approaches seems to have tapered off since there are not that many new features that can be exploited for recognition. It is possible also, that the features currently employed are not able to fully capture the distinct motion of a person and that more distinctive features would require more resources to compute.

The use of full tridimensional models for gait recognition is based on an Analysis-by-Synthesis approach. As we have seen, we need to register the image then reconstruct it. This takes a lot of computing effort and time, not many of the methods described being able to perform in real time. The main problem is that of matching the tridimensional pose with the two-dimensional images, in particular having to generate many tridimensional models so that this task can be done. It is necessary to invoke kinematic and human limb constraints to limit the search for possible tridimensional model movements. This may also be coupled with constraints on smoothness of motion. However, self-occlusion and motion singularities create problems. Recent approaches cast the reconstruction problem as a probabilistic one. The range of actual human limb motions place constraints on the possible positions of these limbs and by assigning probabilities to these positions, an *a priori* distribution is created. Together with the constant stream of image data from video, we update these distributions and thus generate better *a posteriori* distributions of the limb positions.

Most gait analyses are done on a fronto-parallel (FP) plane to a camera. Not many treat the fronto-normal (FN) plane specifically. Gait movement in the fronto-normal plane is treated as a special case of general gait movement and caters for it by the appropriate tridimensional to two-dimensional image transforms. This is done in order to take advantage of being able to observe gait from a distance of 10 meters or more.

As far as experimental considerations are concerned, gait recognition is a relatively new field. Reports of successes have been mainly on small datasets and in ideal circumstances where only a single subject is present. Only recently has there been a standardized test procedure [68] on more realistic scenarios. We survey state-of-the-art gait datasets in Section 5.

## 3. Gait tracking and factors affecting the gait recognition process

Human vision groups raw pixels containing just intensity or colour information into features like edges, corners and regions. As postulated by Marr [69], these features go into initial grouping of entities or what we may call *early vision*. Finally, humans interpret the groups as objects and the relationships between them.

In similar fashion, we use basic features to identify and track gait in video. This is done by finding the temporal correspondence, between the target regions in successive frames of a video stream. Identifying and tracking are closely related because tracking usually starts with identifying gait. So one can say that tracking is actually a series of detections of gait. However, tracking would only need to concentrate on a smaller region of the frame once we locate the gait. We assume that the human subject does not move much between frames.

There are two main sources of video image information that can be used to track gait. First are image features such as color, texture and shape which may be analyzed on a frame basis. Secondly, inter-frame information provides motion cues. After the information is obtained, we may use a combination of statistical approaches using image features and combine it with motion information to give more robust results. The approaches used in the literature are detailed in the next section.

Even with proper tracking, there are a number of spatiotemporal factors to consider in gait recognition, noting that a series of images are needed which change over time. First is the variation in walking speed and the type of clothing and loads carried. Then for views that are perpendicular to the camera plane, the next two factors are quite severe. Looming describes the changes in image size due to perspective projection as subject position varies with a camera. In occlusion, body parts block the view of other parts and these effects need to be mitigated by compensatory measures.

### 3.1. Gait tracking using motion information

Motion-based techniques for tracking rely on robust methods for grouping visual information over time. These methods are relatively fast but have considerable difficulties with non-rigid movements. Tracking is performed through analyzing geometrical or region-based properties of the tracked gait. Depending on the information source, existing approaches can be classified into region-based and boundary-based approaches.

#### *3.1.1. Region-based methods*

These approaches rely on information provided by the entire region of interest such as texture and color as well as motion-based properties using motion estimation and/or segmentation. In this case, the estimation of the target's velocity is based on the correspondence between the related target regions between frames. This is usually time-consuming as a point-to-point correspondence needs to be established in the whole region. However parametric motion models can be used which describe the target motion with a small set of parameters, for example rotation with translation. For non-rigid movements such as gait movements, there may be some difficulty in tracking the actual non-rigid gait boundaries. However we have a more robust result because we use information provided by the whole region and noise may be minimized by least-squares methods.

Methods for doing this start from simple pixel colour grouping using morphological operations. Pixel grouping is fast and may be implemented in hardware. The challenges, as before, are to adapt to changing conditions such as lighting and pose. One of the more recent algorithms introduced by Bradski [70] is the Continously Adaptive MeanSHIFT (CAMSHIFT) based on the Meanshift algorithm. This was given a thorough probabilistic foundation by Comanciu et al [71]. CAMSHIFT allows for dynamic update of colour information in a probabilistic way. Most implementations work at frame rates. However, if color information is not available, for example in gray scale images, a more general method such as optical flow is needed.

In this regard, optical flow [72] is one of the widely used methods for gait estimation. This method is slow as the apparent velocity (vector) of every pixel in the frame has to be known. Given the motion, this can be used to track the gait. Alternatively, if the movements between frames are too large, feature matching across frames can be effected by noting their correspondences [73]. In [37], optical flow is used for tracking in each half gait cycle. Bashir et al. [74] partition flow into four parts in accordance with the direction and symbol, and use the weighted sum of these parts for gait recognition.

#### *3.1.2. Boundary-based approaches*

If no color or texture information are available then boundary-based features (edges) provide reliable information which does not depend on the motion type or gait shape. Boundary based tracking methods employ active contour models, like snakes or geodesic active contours which are described soon. These models use energy or geometry and try to minimize some related quantity. They warp an initial contour under the influence of external forces, while at the same time being constrained by factors such as internal energies or even to follow the contours of a region.

We need to consider that often, we are looking at the two-dimensional projection of a human in tridimensional motion. Thus there will be a deformation of that shape. Technically, this is termed an affine transformation.

Deriving the affine transformation is a standard problem and most often, snakes are used for this. But if the shape is simple, the problem can be simplified.

### Standard shapes

At the basic level, simple shapes like circles and ellipses [75] are used to approximate the shape of the object. As each new frame is received the shape is updated in terms of its parameters, namely the minor and major axes length and position. This may be done using pixel intensity gradients along and/or normal to the contour. From this, the shape parameters are updated, together with the position.

### Snakes and contours

Snakes [76] are used to model more general kinds of shapes. They are defined as deformable active contours used for boundary gait tracking. They move under the influence of image intensity "forces" subject to certain internal constraints on the kinds of deformation possible. So a contour is allowed to "hug" the desired object, based on a lowest energy cost. These forces are related to the gradient of image intensity and the positions of image features. An advantage of the snake model is that it can easily take into account the dynamics derived from time varying images. Snakes are usually parameterized and they are constrained to a predefined shape. So these methods require an accurate initialization step since the initial contour converges iteratively toward the solution of a partial differential equation. However, their computation is time consuming.

Active contour tracking is an evolutionary step from using snakes to track gait.

Much effort has been done to solve the equations of motion and improve performance in the presence of clutter and occlusions. Curwen and Blake [77] use a B-spline representation of active contours, while others employ polygons for tracking [78] and deformable superquadric models for modeling of shape and motion of tridimensional non-rigid objects [79]. As the object moves, we need to update the parameters describing it. Thus the contour is "active".

Geodesic active contour models are not parameterized and can be used to track gait motion. For example [80], a three step approach is proposed which start by detecting the contours of the entity to be tracked. An estimation of the velocity vector field along the detected contours is then performed. At this step, very unstable measurements can be obtained. Following this, a partial differential equation is designed to move the contours to the boundary of the moving entities. These contours are then used as initial estimates of the contours in the next image and the process iterates. In [81], geodesic active contour models are combined with mean-shift algorithm to extract and track human shapes.

From this we see the need to update the parameters of the contours. This is done by trying to find the contour that best fits the shape in the *next* video frame. There are several approaches again, to doing this, using the Condensation algorithm and Multiple Hypothesis Tracking.

### Built-in Motion Information

Recent video standards - for example, MPEG use motion estimation and compensation as as part of their encoding process. This calculates the displacements of consecutive matching blocks of regions in image frames as motion vectors. These are then encoded into standard MPEG streams. In many cases, the motion vector values represent the gait movements very well. It is possible to use these motion vector values directly [82]. There is less accuracy of the boundary as the motion vectors represent movements of a block of pixels. But the availability of such motion information provided "freely" by the encoding hardware makes tracking algorithms in this domain an attractive proposition.

## 3.2. Subject variations in images captured from video

We have alluded to earlier that gait is difficult to disguise, so we postulate that people who exhibit gait traits that are out of the ordinary, for example, walking too quickly or slowly would draw attention to themselves. In light of this we will briefly describe work on several effects that have a covariate effect on human gait such as clothing, footwear type, surface type, load carrying, viewing angle and speed of walking among other effects.

As to the external effects of gait on a subject such as Yu et al. (131) showed only a 1% reduction in recognition rate with variation in clothing and load carrying. In another paper, they (135) demonstrate that a fronto-parallel view is reliatively robust for recognition with varying views. However, in changing the view angle, various forms of normalization and view-invariant features can bring about a certain degree of tolerance in gait recognition.

However large variations in the speed of walking affects recognition profoundly. Even if time independent features were found, the human stride length varies directly with the speed and this also affects the change in height of a person, causing a larger bobbing effect at higher walking speeds. These will all change the silhouettes traced by a person walking, so this is an important consideration.

Some earlier works such as by Tanawongsuwan and Bobick (132) normalize the silhouettes of a human walk according to the stride length. Investigation into compensation for large variations in gait has only recently been addressed for example, Kusakunniran et al. (133) use higher order shape descriptors such as tangents and curvature of a silhouette that are not greatly affected by walking speed. When computing shape moments, they also assign differing weights to parts of the silhouette according to the speed of movement. These measures improve the recognition rate when used in the OU-ISIR A dataset (described in Section 5.2), specifically set up to research into changes in walking speed. Another specialized gait dataset was recently set up to study the effect of speed on gait recognition (134).

## 3.3. Handling looming

In the quest for view-invariant gait feature derivation, objects further from the camera appear shorter than they are - this effect, called foreshortening, needs to be overcome.

### *3.3.1. Adjusting for looming motion*

Some schemes for motion compensation include that of Johnson and Bobick [83] who calibrate a hyperbola modeling the variation in scale of a walker approaching a camera at 45 degrees to the normal. Chowdhury et al. [84] use the direction of the line a subject walks and a calibration object to project the image to a canonical view. Benabdelkader et al. [85] use the edge of the walking silhouette's bounding box to estimate the distance to a calibrated camera and thereby its other tridimensional dimensions. They also compensate for the bobbing movement of a silhouette by fitting the dimensions to a sine wave. Most other approaches use the current dimensions of the bounding box or the silhouette to normalize the related measurements on a frame by frame basis. For example, Wang et al. [86] use the L1 norm of the "unwrapped" binary silhouette as a normalization constant for each video frame as described in Section 2.2.2 and Fig. 3. Another approach proposed by Hu et al. [87] consists of a multi-view gait recognition technique by means of multi-view population Hidden Markov Models.

These methods do not need any calibration or knowledge about the dimensions of the scene, which is a very useful simplification for setup. However, as far as we know, there have not been comparisons done between these methods of distance normalization. For our purposes, we will term these two the "Motion based normalization" and the "Distance based normalization" schemes.

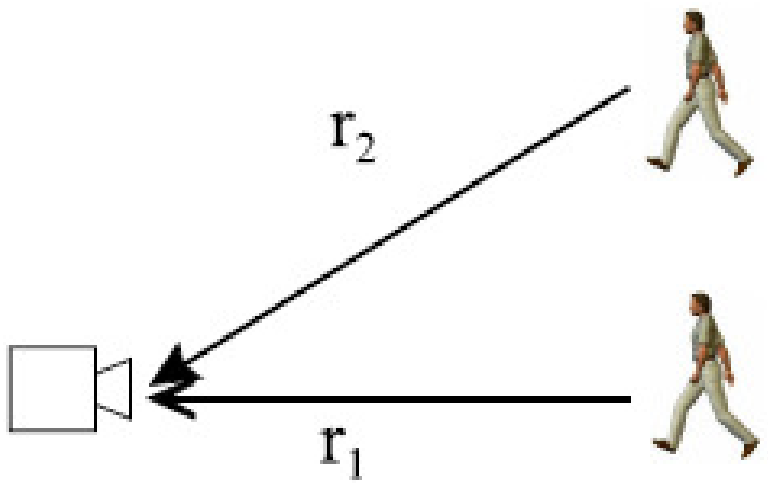

Fig.4: Time to contact vs Range

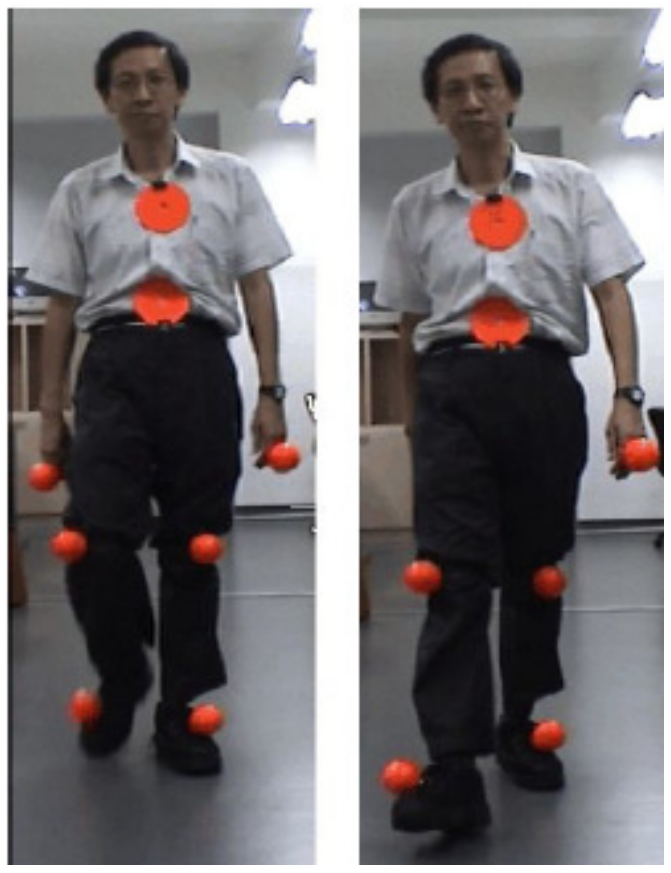

Fig.5: Left coloured marker on hand being occluded

### *3.3.2. Normalization for Looming motion*

The questions of dimension normalization have appeared in the computer vision literature in another form, as the Time to Collision (TTC) problem. The looming of the subject to a camera has many applications. Other terms have been used, like time to contact, time to impact, as well. The applications include detection of collision in robotic and autonomous motion. After all, in order to navigate a course, obstacles have to be avoided. Thus the amount of research in this area is both wide and deep.

Similar to the methods used for normalization, we need to measure the dimensions of an object in an image as it looms closer to the camera. So given the change in image size and the standard lens equation, the absolute or relative distance between camera and subject can be computed. This can be used to compute the time taken for the subject to reach the subject, if we keep track of the speed. In our FN gait setup, this can be done in two ways:

i) Directly, the distance may be computed by considering the relative velocity of the object and the time interval.

ii) Indirectly, if one assumes the complete object is in a plane perpendicular to the optical axis, the change in scale of a feature that has a known, fixed size may be used to scale other features. The distance of the object from the camera does not need to be known as we work with ratios of sizes.

While looming and the TTC are related, Raviv [88] differentiates TTC as the time taken for an object on a plane parallel to the camera axis to reach the camera, while looming considers the range, or actual distance from the camera to the object as shown in Fig. 4. He further quantifies looming as the time rate of change over the relative range. Furthermore, looming can be measured by changes in image texture, area, brightness and degree of blurriness (caused by the image being out of focus).

The methods for computing TTC can be divided into two main areas. Passive ranging systems use tridimensional equations of motion to compute the object speed. The general motion can be constrained, as described by Colombo [89] by either direct translation towards the camera, fronto-parallel motion (pure rotation) or fixating on a small area of the object and using optical flow techniques to compute the divergence of the image field. From the speed, the camera-object distance can be updated. However, optical flow methods are noisy and thus require smoothing of the data adding to the already significant computation burden.

Simpler methods assume constant speed over a period of time. Instead of using optic flow, a rigid image feature is used. By measuring the change in size of this feature, the momentary TTC can be computed. One such approach employing scale-change was used by Dagan et al. [90].

### 3.4. Compensating for occlusion

As we see from Fig. 5 the left hand completely disappears from the frontal view of the camera just a few moments from full view. Of course depending on the type of clothing, other parts may be occluded as well.

Occlusion refers to the blocking of view to an object, implying that the object's movement is being observed. The object was viewable and an obstruction happens. This may be due either to a part of the same object (self-occlusion) blocking the view or another object coming between the line of sight. This problem occurs more often in monocular systems. In multi-ocular systems, a different camera can view the occluded part so there is less chance for self occlusion. However having to combine the inputs of several cameras requires calibration procedures and tridimensional geometry.

In the overview of occlusion tracking by Gabriel et al. [91] most approaches for handling occlusion are mainly interested in when objects merge and split (MS). In our case, we need to use the straight-through (ST) approach for occlusion handling as we need to ascertain the position of the body part at every video frame since we are using the time series generated by gait data for characterization purposes. When an object is occluded, we have to estimate its motion. Some approaches to occlusion treat it as an extension of the tracking problem. Thus, using Kalman filters or particle filters, a probabilistic model of the motion is created.

In motion capture systems, the objective is to analyze human motion with a view to synthesis. In order to reliably compensate for occlusion, Liu et al. [92] use a training set of representative motions with motion markers and build a global linear model of motion. Using existing markers in a video, they predict the position of missing markers using this model.

Other approaches use triangulations from other visible objects which have a fixed inter-object distance and are assumed to be constant as with Aristidou et al. [93] who also use Kalman filtering to predict motion.

Koller et al. [94] use depth information to determine the order of occluding objects (cars) to remove unwanted image data. Kalman filters are used to estimate the object contour and its motion even in cases of occlusion.

Blake et al. [95] use particle filtering with partitioned sampling to track multiple objects with various degrees of occlusion.

However, all these efforts at tracking cannot be sustained for long if the object is completely occluded or the motion is complex.

The problem of occlusion may be recast in a different form if we consider that the points at which the object is occluded are "missing" from the main set of data. A large body of work exists in the DSP (Digital Signal

Processing) literature which considers how to reconstruct missing or corrupted data which become outliers. A variation on this is the reconstruction of a signal based on irregular sampling. The missing samples may be considered as making the signals irregularly or non-uniformly sampled.

The theory for signal reconstruction from regularly sampled data is well developed, judging from standard texts on DSP. However judging from these texts, the dearth of work on reconstruction from *irregular* samples indicates that this is a specialized field.

Depending on how one views the problem, this approach can be used for interpolation and extrapolation. In fact, the applications extend to superresolution [96] of signals, as the "missing" samples can be considered to be the higher resolution data of which only the known samples are available. The concepts can be also be used in fault tolerant computing, to detect the presence of erroneous data by *predicting* what is the correct data and comparing with what is actually present. The concepts have also been used in astronomy [97], to compensate for missing data transmitted by satellite, medical imaging, to fill in gaps when data could not be acquired.

From the 1980's the early on signal reconstruction started off with ad hoc methods based on geometrical considerations, using splines to bridge the gap between unevenly spaced samples. Splines, by their smooth spatial qualities are a natural choice for this. This then led to prediction methods using Autoregressive (AR) modelling for example. Esquef et al. [98] use AR modelling to predict the gap in the data due to missing samples. As AR modeling is so often used for prediction purposes, here it is used to extrapolate from previous samples. However, if the gap is large, the AR prediction becomes less effective and a *backward* prediction needs to be done from samples after the gap.

The earliest example of using both time and frequency domain manipulation of data to obtain missing samples is the Papoulis-Gerchberg [96] method. The data is alternately converted between the time and frequency domains to reconstruct a time domain signal in a typical projection onto convex sets (POCS) setting. In the frequency domain, the signal is bandlimited whereas in the time domain, the *original* (existing, without missing) data overwrites that generated from the frequency domain information. A useful description can be found in the thesis of Souppa [99]. The steps are:

i) Fourier Transform the signal.

ii) Band-limit the Fourier transformed signal.

iii) Inverse Fourier Transform the bandlimited signal to get a reconstructed signal.

iv) Replace the portions of the reconstructed signal that were *not missing* from step i)

The above steps are repeated until the mean-square error between the reconstructed and original signal is below a threshold. In the literature there has been a large number of variations on this theme, at times applied to two-dimensional signals (images) as well.

However, these use the entire sampling set, of both known and unknown data which can be substantial. A great savings in computation can be achieved if a subset of the data can be used. The so-called Minimum Dimension formulations by Feichtinger et al. [100] were done in the frequency domain. Here they show that if an irregularly sampled signal is bandlimited, one needs to solve for only the number of unknown samples. Ferreira devised a time domain version [101] of the Minimum Dimension formulation. He generalizes further on this [102] by allowing the choice of frequency or time domain approaches, depending on whether there are more unknowns when formulated in either domain.

It should be noted that most algorithms started off analysing a continuous signal. However, when considering discrete signals, questions of stability come in as faster methods for solving the relevant equations are sought. The ability to cast the problems into a form using Toeplitz matrices accounts for the speed of these algorithms. This is explained in Feichtinger's work which was mentioned in the previous paragraph.

A generalization of the set of Fourier functions used leads to concepts involving Hilbert space from which theory of frames evolved. Frames provide an elegant and comprehensive mathematical base from which to compute numerical aspects of the algorithm, like convergence, error estimates, and stability. There is also the possibility of using more general kinds of functions. Frames lead to wavelets and filter bank approaches to irregular sampling.

An initial application was by Eldar et al. [103]. Later approaches which consist in recovering periodically spaced missing samples in signals can be found in works like Bariska's [104].

## 4. Combination of gait with other biometrics

In the fusion of biometrics, Jain et al. [105] summarize the state of art by describing several combinations e.g. face and speech, face and iris and so on. The combination of biometrics takes place at several levels. In hierarchical systems, a less discriminating biometrics can prune off unlikely candidates from a dataset. We allow a higher rate of false positives to be passed to other biometrics and use features that do not require many resources to calculate. This leads to a reduction in search space and thus gives a faster speed in matching. Of course if one considers the rate of success in classification, it will go up as the system works on a more likely set of candidates, and thereby compensates for the loss in accuracy, if any. In holistic systems all available data is used and combined in various ways with various rules to give an overall result [106].

Gait is a biometric that can be measured from a distance, far before a face can be clearly seen. Thus it can be used to pre-select a group of possible candidates to check. When the person draws near enough for face or even iris recognition to take place, the number of people to check can be drastically reduced, thus giving an overall better recognition rate.

Thus a promising area in recognition is the fusing of several biometrics to make the overall process more robust. We need to note that gait seems to work best in the fronto-parallel plane. In conjunction with face features and the iris, these work with the subject facing the camera, which would seem to necessitate the use of two cameras. It is notable that the combination of face and gait biometrics is of sufficient interest that it has warranted the setting up of specific datasets to research as we shall see in Section 5.3. This may be due to the non-invasive nature of the images captured, and that from a distance as well. The use of face as a biometric was established by Turk and Pentland in their seminal paper [136] and this field of research is still very active. This has led to several recognition systems in use, incorporating face as a biometric. Face recognition (FR) works with images in a bidimensional realm. With the availability of hardware that makes 3D capture feasible then, paralleling the approaches in gait research, there is now work in FR in a tridimensional plane [137] [138].

Liu and Sarkar [109] use EBGM for face recognition and stance based gait recognition. The stances are derived from silhouettes of a walker. They use Gaussian z-normalization scores separately from face and gait recognition. They then use a variety of fusion measures like sum, Bayesian, weighted score and ranking of the sums from the two biometrics. In all cases, fusion had a beneficial effect.

Most of these methods require the use of two cameras. However, some novel monocular approaches are those by Zhou and Bhanu [110] who use a *profile* view of a face with gait in order to use one camera at 3.3 m from the subject. For even more biometrics, the work by Bazin [111] includes the ear and footfall as well.

## 5. Survey of experimental datasets involving gait

In this section, we survey three types of datasets that incorporate gait. First are the gait datasets used for recognition. Then we look at those that combine gait and face and finally the motion capture datasets.

Table I. Motion capture datasets with walking

| Name/ Sponsor | Hard ware | Resolution/ Rate | Subjects | # cams | Actions |
|---|---|---|---|---|---|
| HumanEva-1 Brown Uni. | Vicon/ Pulnix | Vicon-120 Puln-VGA/110&60 | 4 | 7 video 5 mocap | walking, jogging, gesturing |
| CMU Graphics Lab Carnegie Mellon Uni. | Vicon | n.a/120 | 31 | 12 | walking - other actions available |
| ACCAD Ohio State Uni | Vicon | n.a/120 | 7 | 14 | walking - other actions available |
| HDM05 Uni. Bonn | Vicon | n.a/120(240) | 5 | 6-12 | walking - other actions available |

Table II. Current available gait datasets

| Name/ Sponsor | View | Other views | Resolution/ Rate | Subjects/seq/ cycles | # cams | Platform | Height |
|---|---|---|---|---|---|---|---|
| GA Tech/2001[113] | FP | 45/135 | QVGA/30 | 20/194 | 3 | markers | |
| UCSD/1998 | FP | n.a | 480x640/ | 6/42/3 | 1 | indoor | ~4.5' |
| CMU Mobo www.hid.ri.cmu.edu | FP/N | Multi-angle | VGA/30 | 25/16 | 6 | treadmill | 30' |
| MIT/2001 | FP | n.a. | 720x480/15 | 24/225/3 | 1 | indoor | 4.5' |
| USF/2001[114] | FP/N | ellipse | 720x480/30 | 122/1870 | 1 | in/out/load | ~4.5' |
| UMD/2001 | FP/N | to/fro | QVGA/20<br>Phillips G3 Envirodome | 25/100 | 2 | in/outdoor | 15' |
| UMD/2001[115] | FP/N | T- shape | VGA/20[i]<br>150x75/20 | 55/220 | 2 | outdoor | 15' |
| UMD3/2004[116] | FP | 15/30/45/60 | 210x105/25 | 12 | 2 | indoor | 4.5' |
| CASIA A/2001 | FP/N | 45 | QVGA/25 | 20/12/3 | 2 | in/outdoor | ~4.5' |
| CASIA B/2005 [117] | FP/N | various | QVGA/25 | 124/12/3 | 11 | in/outdoor | ~4.5' |
| CASIA C/2005 | FP/N | various | QVGA/25 | 153/12/3 | 11 | IR | ~4.5' |
| Soton[118]<br>U Southampton | FP | n.a. | 384x288/25 | 6/4 | 1 | indoor | 4.5' |
| SotonBig/2001 | FP | oblique | 720x576/25 | 115/2128/1.5 | 2 | in/out/tread | 3’ |
| Boyd[119]<br>U. Calgary | FP | n.a. | 320x160/30 | 6/22 | 1 | indoor | ~4.5' |
| LAB5(2008) *<br>Uni. Liege | FN | n.a. | VGA/25 | 5/4 | 1 | indoor | high |
| LAB21 *<br>Uni. Liege | FN | n.a. | VGA/25 | 21/4 | 1 | indoor | high |
| OU-ISIR A[133]<br>Osaka Uni. | FP | Various | VGA/60 | 34 | 25 | treadmill | 4.5’ |
| OU-ISIR B | FP | Various | VGA/60 | 68 | 25 | treadmill | 4.5’ |
| OU-ISIR D | FP | | VGA/60 | 185 | 25 | treadmill | 4.5’ |
| OU-ISIR A | FP | | | | | | |

* - not available publicly

### 5.1. Motion capture datasets

We consider motion capture datasets in Table I for an idea of the hardware and software requirements. As the motion is more important than the identity, the number of subjects is less, but of course the number of different human actions recorded is higher but all include the action of walking.

The datasets are publicly available, with authorization needed. All the sets use the Vicon tracking system which produces the coordinates of markers attached to the body, sometimes up to about 40 attached markers.

### 5.2. Gait only datasets

A survey of currently available gait datasets is shown in Table II. A good current description of the sources of biometric datasets is available [112]. Nevertheless a countercheck showed variations in the information. This is probably due to the dataset publishers updating the contents. Also where data is not available, only when the actual video clip is examined can the capture parameters be seen. Relevant, extra information not readily available is attached as comment.

We make the following observations:

i) All the datasets are unimodal, i.e. they focus on one main biometric which is gait.

ii) Few use currently available DV technology at Standard Definition (SD) resolution which is 720x576 (PAL) or 720x480 (NTSC). This could be due to the fact that mostly silhouette analysis is used, which does not need the high resolution.

iii) Frame rates range from 15 to 30 fps.

iv) Most videos employ lossy encoding (DV, webcam).

v) Most common viewing angles are FP.

vi) Most cameras are placed at half body height.

vii) Number of videos recorded were 122 on the high end, to about 20 for a medium sized dataset.

viii) Lately, exclusively FN gait databases have appeared, for example from University of Liege.

### 5.3. Gait and face datasets

We next consider datasets used in conjunction with face recognition provided in Table III. The number of datasets mentioned in the publications is quite small. The number of subjects participating is quite small as well.

The following observations are made:

i) The CASIA A dataset was originally used for gait recognition only. A member of the team, Wang, who spearheaded the use of the original CASIA datasets, reused the dataset for face as well in the current application by Geng et al [106]. Due to the low resolution, the face images were normalized to 25×25 pixels and the body silhouette to 48×32 pixels.

ii) University of California at Riverside pioneered the use of face profiles, i.e. from the side, for identification. From these, we note that there is only one dataset publicly available currently for the combination of face and gait recognition. Except for UCR B, the number of subjects used is in the low tens. Presumably these are early versions of the datasets to determine the effort required for larger scale experiments.

As for the suitability of the images for face recognition, one of the largest publicly available dataset, the FERET [120] for face recognition uses a distance of 40 to 60 pixels between the eyes, implying a 10% or so larger area for the face. Czyz et al. [121] determined that even down to 16 x 16 pixels, automated recognition still performed acceptably. Hence we conclude that using the CASIA A dataset offers a resolution at the lower end for reliable face recognition. In contrast, with DV (720x480 pixels) resolution, we report later that the region of interest containing the face ranged in size from 22x26 to 75x91 pixels as the person nears the camera.

With commercially available HD video recorders operating at vertical resolutions of 1080 pixels (twice the previous generation) at a reasonable cost, there is no reason to graduate to higher resolutions for algorithms to work with.

Table III. Current gait datasets used with face recognition

| Name/<br>Users | View | Other views | Resolution/<br>Rate | Subjects/seq/<br>cycles | # cams | Face image size |
|---|---|---|---|---|---|---|
| MIT/2001 *<br>Shakhnarovich | FP/N | 45° apart | QVGA/30 | 12/2-6/(3 m) | 4 | n.a. |
| CASIA A/2001<br>Geng | FP/N | 0,45,90 | QVGA/25 | 20/12/3 | 1 | 25x25 |
| UCR A *<br>Zhou & Bhanu | FP | n.a. | 480x720 | 14/2/various | 1 | side view |
| UCR B *<br>Zhou & Bhanu | FP | n.a. | 480x720 | 46/2/3 | 1 | side view |
| NIST 2002 [122] **<br>Kale et al. | FP/N | Σshape | | 30 | 1 | n.a. |
| USF 2001<br>Liu et al. | FP/N<br>*separate* | Face-FN<br>Gait-FP | 720x480/30 | 70 | 2 | |

* - not available publicly ** - superseded, not available

### 5.4. Conclusions on gait datasets

A cursory glance at the available datasets shows that analyzing gait to identify a person from multiple viewpoints is useful only for *uncluttered* environments. Most datasets focus on the FP view for example in the USF dataset, even though the subject walks in an elliptical shape, the subject spends most of the time in the FP view.

To employ multimodal biometrics using the FN viewpoint requires a dataset with higher resolution for the near-distance biometric to work. Current datasets have relatively low resolution of the region of interest - namely the body shape and the face.

The mocap datasets sample at 120 Hz as compared to 30 Hz for the gait datasets. This shows that for finer analysis of motion, we need a higher sampling rate of the motion. Finally, we note that most of the early versions or early datasets had a smaller number of subjects presumably to establish operating parameters for large scale capture of gait data.

## 6. Perspectives for the use of fronto-normal gait

In the main, current gait recognition approaches analyze walking that goes on in a plane parallel to a camera, the so-called fronto-parallel view. This gives the largest variation in a silhouette as the legs and hands extend to their maximum distances when walking. From a far distance, this is advantageous. Motion from a plane perpendicular to this, the fronto-normal view, is considered as a special case. The variations in silhouette are smaller, but this view has its own merits. Based on the personal experience and work of the authors [123] and others, the FN view shows highly idiosyncratic movements of the person's feet and finger tips, although smaller in magnitude than the FP view. Together with the height, which is a static measure, our senses can recognize people by these cues alone. Associated with this is the ability to view the subject for a longer period as they approach us, compared to people who walk in a FP plane, where they disappear from view quickly. While it is useful to devise gait recognition algorithms that work in all types of environments, in reality practical considerations constrain the way sensors and cameras work. For instance, a common scenario is where people have to queue up to access a facility, e.g. passing through immigration checkpoints, secured facilities or entering a building as shown in Fig. 6. In situations where there is a need to check for the identity of a person, people line up and due to physical constraints, as they walk towards the point of access where a camera may be placed. So instead of requiring a large, clear field of view, the physical space needed is reduced and onr can employ a far-distance biometric like gait and combine it with a near-distance biometric as face in a natural, practical and useful way. We motivate for the use of FN gait from the following considerations.

### 6.1. Space Constraints

Commonly, people are made to queue up to access a facility. In a corridor like structure, we assume that a subject approaches a camera. Depending on the type of analysis need, in a FP walk, at least two cycles or four steps are needed. For more robust estimation of the period of walking, about 8 m is recommended [124]. To capture this movement, the camera distance required is about 9 m [125]. This is because current video cameras typically have a focal length and sensor size of 8 mm and 1/2" respectively. Practically, having such a wide uncluttered space is difficult, since whenever we want to measure a person's gait, many people and objects will be present. We also note that in order to use a combination of biometrics, two cameras are usually needed.

In a FN view, a distance of eight meters can still be used. But this time, twelve steps are covered and there is only the need for a corridor-like structure, the width being about that of a human body. Therefore, considerable amount of space is saved as shown in Fig. 7, in this case by 2/9. Another way of looking at this is that for the same amount of space, the FN view provides a longer period to observe a person's behaviour which is always useful.

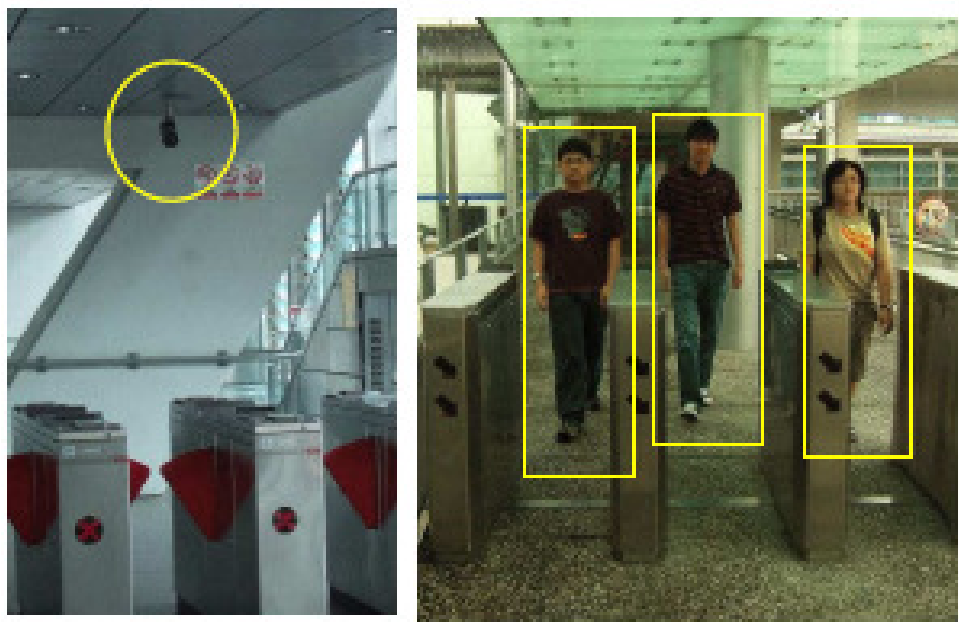

Fig. 6 Left, view of typical security camera monitoring access point
Right, tracking multiple gaits

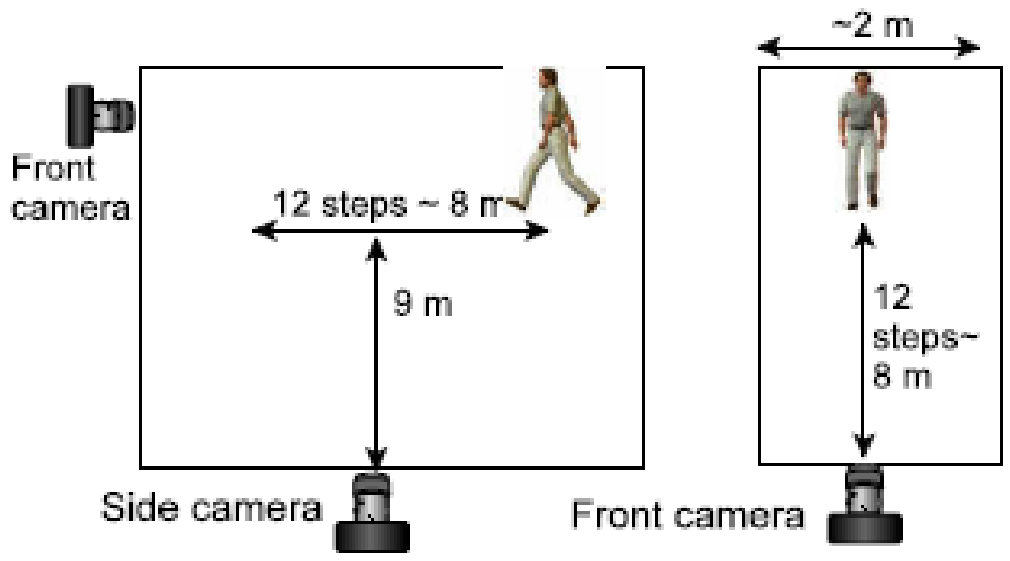

Fig. 7 FP vs FN – physical dimensions for video capture

### 6.2. Coupling the FN characterization of gait with face

In most cases where a framework combining face with gait has been explored, two or more cameras are needed where problems of alignment and synchronization are significant. A single camera or monocular capture of video is preferred. From Table IV, we see that the FN view allows one to use face, iris, and gait for a robust recognition system.

Table IV. Types of biometric combinations possible with the two views of gait in a monocular set up

| Biometric | FP (side view) | FN |
|---|---|---|
| Face | Not reliable | Frontal - well researched |
| Gait | Good segmentation strong periodicity | Difficult to process Can use nonlinear |
| Iris | Not possible | Near distance use |
| Ear | Not sure of usefulness | dubious use |

### 6.3. Psychophysical and other experiments

The research by Wang et al. [126] show that even FN silhouettes give better recognition performance and Troje [127] has shown that the task of recognizing gender from Moving Light Displays has a better error rate using the FN view. This corresponds to earlier works by gait researchers as mentioned in his paper.

### 6.4. Dynamic information

In earlier papers, Lee et al. [128] show that FN gait can be characterized by nonlinear measures which show potential to be used as a biometric. Also, a variety of time series analyses may be employed further to characterize the gait such as in [129]. In contrast, FP gait yields mainly linear, periodic measures.

### 6.5. Summary of FN gait advantages

In summary, the main advantages of monocular FN non-silhouette approach are:

i) Smaller physical space is needed.

ii) Multiple subjects can be tracked.

iii) Other biometrics can be easily combined.

iv) Wide variety of time analysis including non-periodic motion analysis can be used.

However, the FN view is challenging, having to compensate for the looming effect as well as the possibility of self-occlusion happening, when hands and feet are obscured from view by the body, the opposing hand or foot or clothing.

## 7. Summary, conclusion and perspectives

We have looked at various gait recognition techniques and datasets. Currently, these mainly use a FP view of gait and use the silhouette of a subject for analysis. The silhouette is converted into a one dimensional signal and frequency domain techniques applied to it. Most currently available datasets are geared for this arrangement, the image of a person being of relatively low resolution since they are far from the camera. We also saw that the tridimensional types of approach to gait recognition are computationally expensive and thus used for medical and computer animation applications where analysis run times are not a major problem.

However, from the discussion in Section 6, we saw that the FN view was a more practical way to view a person. It can also be combined with other biometrics easily, allowing for more robust identification. However, there are no publicly available datasets specifically geared for multimodal person identification. In setting up our own dataset, we considered using coloured markers for easy tracking of body parts for a finer movement capture as compared to silhouettes. We saw that colour segmentation using probabilistic methods is fast and robust. In overcoming the problems inherent in FN gait, we have to first adjust for the looming effect and we saw several approaches. In respect to compensating for partial and complete occlusion, we saw that the motion of an object needs to be modeled so it can predict when the object disappears from view. However, a novel view is that this problem can be cast as recovering the "missing" data from a time series.

We then examined approaches to multi biometric recognition. Specifically, we looked at combinations that were amenable to our aim of practicality and usefulness. With this in mind, the scope of future works should be narrowed to where the strengths of gait and face recognition may be employed. That is the fronto-normal situation, which is actually very common in practice. Now, long range surveillance of a subject cannot predict where a subject will move to. Thus gait recognition has to be pose and view-independent in this scenario. It also requires a clear view in order to study the subject.

But in most everyday situations where there is a need to check for the identity of a person, a doorway of some sort is needed. For example in order to enter or exit a building, a queue of people form for this purpose. For example, in queues at positions of entry or exits at immigration checkpoints, secured facilities, humans form up and leave in an orderly manner because of physical constraints. Thus focusing on this combination of biometrics is a very practical and useful approach to the recognition problem.